# Learning Visual Representation of Underwater Acoustic Imagery Using Transformer-Based Style Transfer Method

Xiaoteng Zhou[a], Changli Yu[a*], Shihao Yuan[b], Xin Yuan[a*], Hangchi Yu[a], Citong Luo[a]

[a]School of Ocean Engineering, Harbin Institute of Technology, Weihai, China;
[b]School of Computer Science and Technology, Harbin Institute of Technology, Weihai, China

## ABSTRACT

Underwater automatic target recognition (UATR) has been a challenging research topic in ocean engineering. Although deep learning brings opportunities for target recognition on land and in the air, underwater target recognition techniques based on deep learning have lagged due to sensor performance and the size of trainable data. This letter proposed a framework for learning the visual representation of underwater acoustic imageries, which takes a transformer-based style transfer model as the main body. It could replace the low-level texture features of optical images with the visual features of underwater acoustic imageries while preserving their raw high-level semantic content. The proposed framework could fully use the rich optical image dataset to generate a pseudo-acoustic image dataset and use it as the initial sample to train the underwater acoustic target recognition model. The experiments select the dual-frequency identification sonar (DIDSON) as the underwater acoustic data source and also take fish, the most common marine creature, as the research subject. Experimental results show that the proposed method could generate high-quality and high-fidelity pseudo-acoustic samples, achieve the purpose of acoustic data enhancement and provide support for the underwater acoustic-optical images domain transfer research.

## 1. INTRODUCTION

As the human exploration of the ocean continues, more and more sensors are being developed to acquire rich underwater data. Due to the absorption and refraction of light by the seawater medium, optical cameras are limited in their range of action. The cameras usually need to be equipped with high power-consuming external light source devices, which leads to their limited application space underwater. These limitations are significantly magnified in deep water environments. In contrast, sonar sensors have become the primary sensors in underwater detection tasks due to their long detection distance and freedom from turbid zone interference [1]. With the development of the electronics industry, imaging sonars with high resolution have been successfully developed, such as the dual frequency identification sonar (DIDSON) and adaptive resolution imaging sonar (ARIS), which belong to a class of forward-looking sonars that are similar to cameras due to their DIDSON devices are more flexible than side-scan sonar (SSS) and synthetic aperture sonar (SAS) and can significantly reduce the error caused by carrier platform motion, and also have better imaging results than mechanical scanning sonar. High-performance imaging sonar improves the cost-effectiveness of underwater acoustic vision applications. Further, it accelerates the expansion of acoustic image-based marine engineering applications while laying the foundation for introducing machine learning and deep learning methods. Currently, underwater acoustic images acquired by DIDSON devices are widely used in fisheries, underwater archaeology, pipeline maintenance, and personnel search and rescue [2-4].

Traditional underwater automatic target recognition tasks usually require three parts: pre-processing, feature extraction, and feature classification; however, most of the methods used in these three parts are developed for optical images and are not compelling enough when applied to underwater acoustic scenes. These traditional methods also rely heavily on researchers' a priori knowledge in parameter tuning. A good solution usually requires a lot of trial and error, which is not friendly enough for newcomers in underwater vision. With the rapid development of deep learning in computer vision, more and more target recognition tasks are performed using convolutional neural network (CNN) architectures. The recognition results have been widely validated [5-6]. These learning-based recognition algorithms adopt a data-driven approach. The model learns features and their underlying laws from a large number of labeled training samples and uses them for end-to-end classification and recognition, with the whole process approaching an end-to-end model without complex manual design. Also, deep learning-based tasks for underwater applications require less expert analysis and fine-tuning, and the knowledge and expertise required to process underwater acoustic images manually are replaced by the knowledge and expertise required to iterate using deep learning architectures [7], a process that has changed as shown in Figure 1, which is in line with the trend towards intelligent underwater applications.

*yuchangli@hitwh.edu.cn; xin.yuan@upm.es

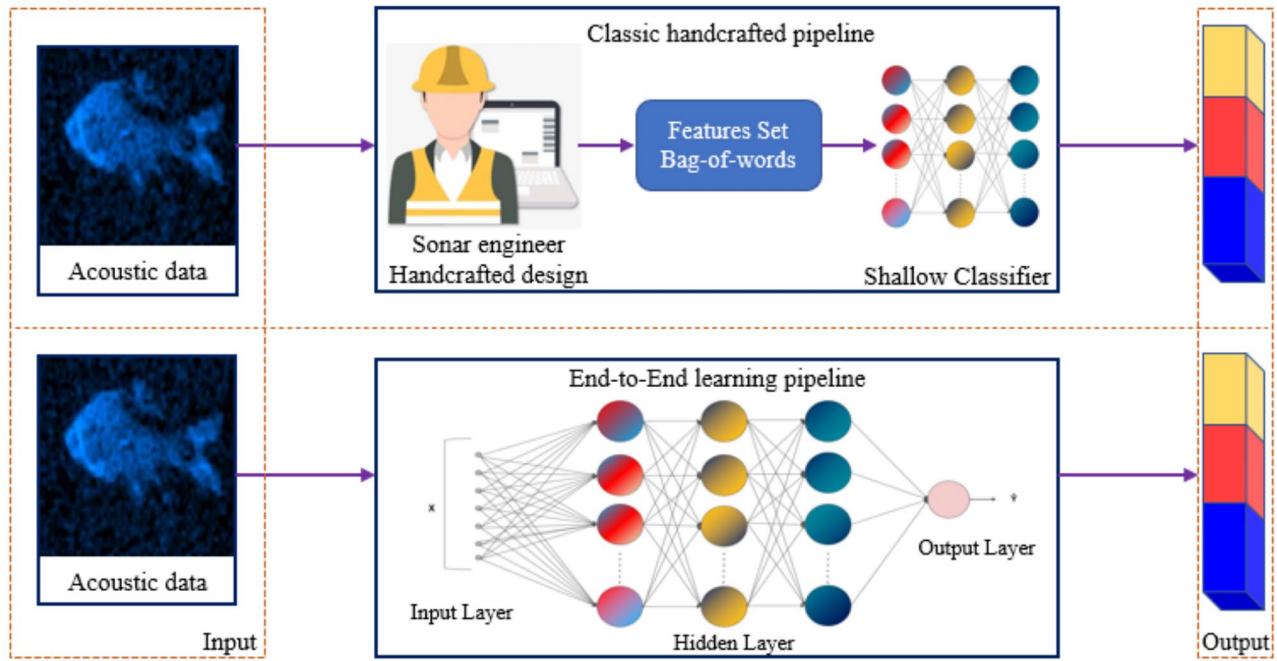

Figure 1. Development comparison schematic of underwater automatic target recognition (UATR) technique.

However, while deep learning brings opportunities to underwater acoustic target recognition, it also poses significant challenges. CNN-based approaches require a large amount of labeled training data upfront, and the features obtained from deep neural network learning are based on a specific training dataset. In short, if there is a problem constructing the data set used for training, the network will become less effective in processing images outside the training data set. However, due to the complexity and unknown nature of the underwater environment, it is expensive and time-consuming to collect large amounts of underwater acoustic data.

In computer vision research, some researchers have suggested the concept of transfer learning [8], which entails fine-tuning a model trained on high-volume non-target data with low-volume target samples to get superior prediction results finally. However, attempts at this idea tend to focus on the same domain, i.e., features where the training and target images have small heterogeneity and commonality, such as two optical imaging domains or two remote sensing imaging domains. Due to the different imaging methods used by optical cameras and imaging sonar in underwater situations, the final images obtained have a substantial domain offset problem [9], which makes transfer learning face a significant challenge in underwater acoustic target recognition tasks. Transfer learning requires a certain amount of initial target data, but it cannot be fully supported because the water environment to be explored is usually harsh and complex, and most underwater targets are zero samples or extremely scarce. For this extreme data scarcity, some researchers have proposed synthesizing pseudo-samples, essentially a kind of data enhancement. Style transfer is a deep enhancement technique instead of data enhancement techniques in the conventional sense, such as rotation, cropping, or adding noise. It is based on the fundamental idea of keeping the high-level semantic information, also known as content, in the image and then replacing low-level texture features, also known as styles [10]. Style transfer-based training data enhancement methods are very flexible, unlike image simulation software-based and GAN network-based methods [11-13], as these methods often require experience from pre-preparation. The pseudo-acoustic sample generation method based on style transfer is more suitable for scenarios where the initial sample is unknown, such as identifying unknown fish in unknown waters. When a specific size of training samples is generated, they could be used to train recognition and classification models for underwater acoustic images, which will be more in-depth and flexible than the transfer learning-based training model.

Using style transfer methods to generate underwater pseudo-acoustic image datasets and thus complete the initial training sample accumulation is effective [14-15]. However, their study was conducted for underwater side-scan and multibeam sonar images. There is a restricted application space for the pseudo-acoustic datasets after style transfer since the images detected by these two sonar sensors are still visually distinct from the optical ones. Unlike them, this paper introduces a database of images detected by DIDSON devices, which are very close to optical images in terms of visual effects, i.e., it is assumed that only domain differences exist between the two. Relying on the progressive nature of underwater acoustic data, the research in this paper directly serves underwater fish identification applications, which has proven to be an urgent need in many studies [16-17]. In addition, the style transfer model used in this paper is implemented based on the transformer principle, a transfer model that introduces a global attention mechanism that has been shown to surpass CNN-based transfer models in a variety of ways [18].

This letter proposed an underwater acoustic image style transfer method to enhance acoustic training datasets. In particular, the sensor used in this study is the acoustic camera DIDSON because its visual imaging effect is most similar to that of an optical camera, and it demonstrates substantial practical value in current mainstream underwater acoustic vision research, serving the marine task of identifying unseen fish in uncharted waters, which is one of the most frequent and complex challenges. One of the core style transfer methods is based on transformer design, which has more evident advantages in global style feature transfer. After generating the pseudo-acoustic dataset is systematically evaluated and analyzed further to assess its value in real marine engineering applications. The overall research framework proposed in this paper is shown in Figure 2.

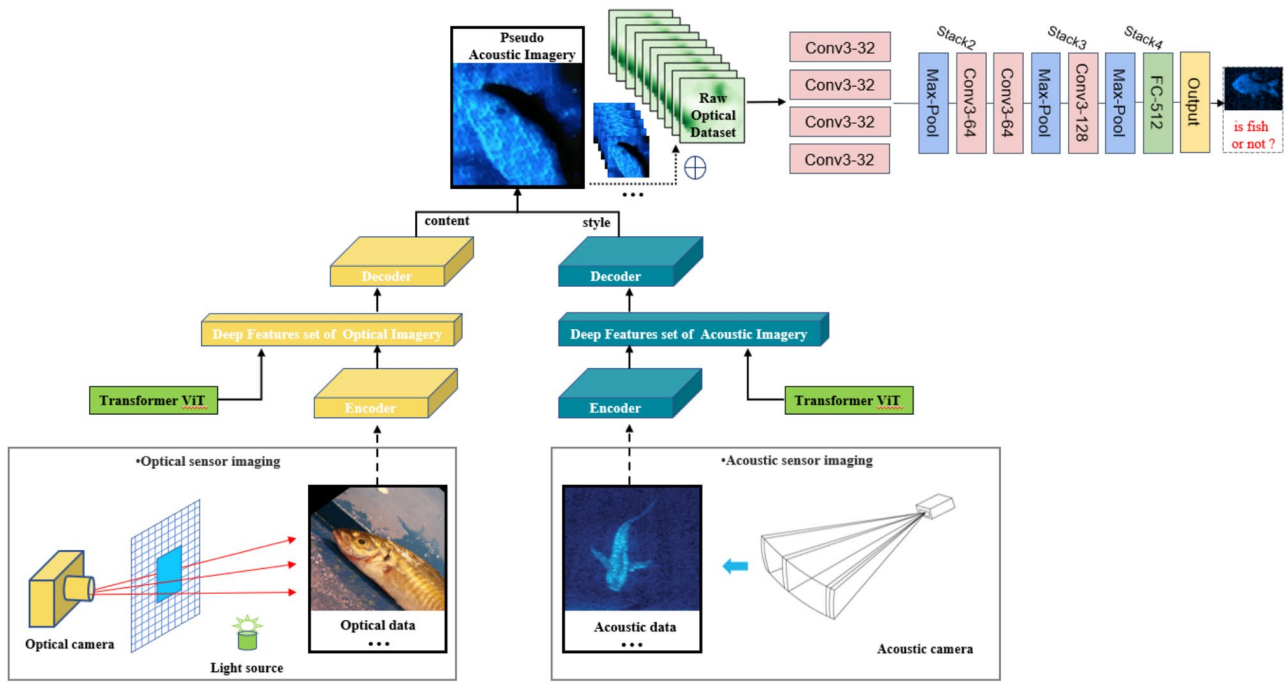

Figure 2. The overall research framework proposed in this letter.

To create underwater acoustic images with incredibly realistic effects, first input the existing optical image samples and the acoustic images of the underwater target scene. Then, using the transformer architecture, extract the high-level semantic content provided by the former and the acoustic imaging style provided by the latter. Finally, wait for their decoding and encoding operations to be finished. A rich pseudo-acoustic dataset can be generated by iterations and loops, which can then be optionally fused with the original optical image dataset for subsequent model training or used alone for model training.

Since most of the current automatic sonar image recognition tasks are implemented on private datasets, few publicly available models are directly used for practical underwater applications, which is one of the motivations for writing this paper. In this paper, the proposed framework is validated using underwater acoustic images captured from natural, unknown marine environments, and the specific contributions are summarized below.

- ◆ A transformer-based style transfer method is proposed to be introduced to underwater acoustic images and is optimally adapted.
- ◆ Generating pseudo-acoustic image datasets using a style transfer-based pipeline and systematically evaluating and analyzing them.
- ◆ The research is based on high-resolution DIDSON data for fish identification tasks in unknown waters.

The remainder of this article is organized as follows. Section II describes the details of the proposed method. Section III illustrates the experimental setup and validation method. Section IV analyzes and evaluates the experimental results. Section V discusses the shortcomings of the experiments and gives an outlook. Conclusions are given in section VI.

## 2. METHODOLOGY

### 2.1 Style transfer model

In order to improve the practical application value of this research in marine engineering, the design of the method in this paper is not limited to a specific link. However, it proposes a complete research framework, aiming to directly connect the resource-rich optical images with the sample-poor acoustic images through this framework, where the optical images provide high-level semantic information to exploit their imaging advantages effectively. In contrast, acoustic images only provide stylized textures to avoid their imaging limitations cleverly. Finally, the generated pseudo-sample dataset is used to train downstream task models, such as underwater target recognition and segmentation models. In the traditional transfer methods, it is frequently necessary to continuously deepen the number of network layers to improve the transfer effect. However, this will result in a reduction in the image's resolution and the loss of feature details, which will produce an incomplete and inaccurate representation of the final migrated image content and prevent the style texture from being adequately embedded. For this problem, the approach of this paper mainly applies the latest research in the field of style transfer using transformer [18] to the field of sonar image generation and compares the gap with other methods. The model architecture developed in this paper has a better global feature transfer effect and more robust feature representation capability, which can minimize detail loss during the image transfer process while ensuring the structural integrity and higher quality of the generated pseudo-acoustic images. The basic idea of a transformer-based style transfer network adopts the vit approach, which is more widely used in computer vision, to improve the performance of the transformer on the style transfer task by proposing the spatially invariant positional encoding method content-aware positional encoding (CAPE), which avoids the sensitivity of transformer to scale changes while ensuring that transformer captures the long-range information of the image.

The transformer used in this model is originally a robust structure in natural language processing. Its basic structure consists of an encoder and decoder, where the encoder converts the sentence and position encoding composed of the token into intermediate encoding through the attention mechanism. Then, the decoder converts the intermediate encoding into output results; thus, the sequence-to-sequence generation task is achieved. Both encoder and decoder have a self-attention mechanism to extract semantic information, and then the decoder's attention mechanism to translate. The transformer can also perform the image-to-image generation task by decomposing the image into several patches and then encoding the position. By introducing CAPE, the structural information of the content image is preserved. In contrast, the traditional sin position coding of the style image loses the structural information so that only the style texture information is preserved. Then the intermediate coding of the two parts is mixed and decoded by the transformer's decoder to form the image after style transfer.

This letter applies the transformer-based style transfer method to pseudo-sonar image generation. Since the style transfer algorithm can only migrate style texture information, we perform image enhancement on the optical image prior to transfer to optimize it for the pseudo-sonar image generation task. The leading optimization considers that the sonar image is a single-channel colorized pseudo-color image, so the original optical image is grayed out to simulate a single-channel acoustic image. Also, considering that the background of the sonar image is usually featureless seawater, while our optical image is an especially captured fish image, we perform foreground extraction for the optical image to simulate the effect of an actual sonar image without a background. Here, the foreground extraction is performed using the Image Matting method, and a manual assistance method is used to make the effect more accurate. In addition, the necessary image processing, such as Gaussian smoothing, is also used appropriately to make the result of optical image transfer closer to the actual sonar image. The overall image processing pipeline is shown in Figure 3.

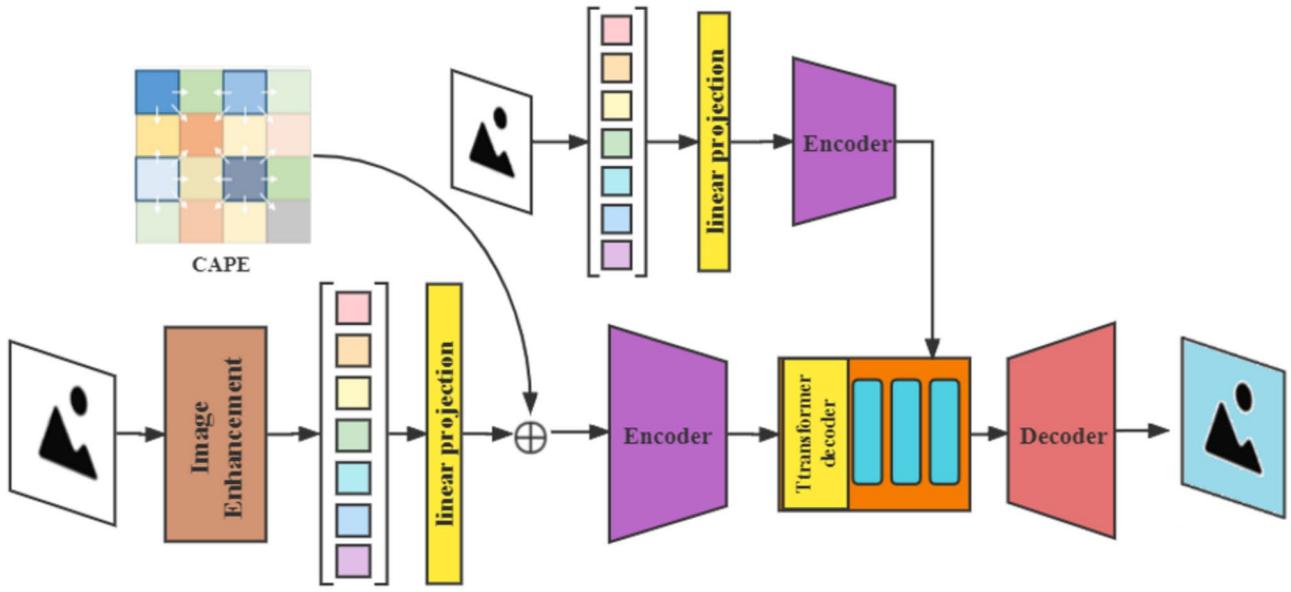

Figure 3. Schematic diagram of the overall image processing pipeline.

## 3. EXPERIMENTAL SETTINGS

This subsection describes the details of the experimental setup, the source of the training dataset, and the strategy for validating the experiments, and the overall architecture is shown in Figure 4.

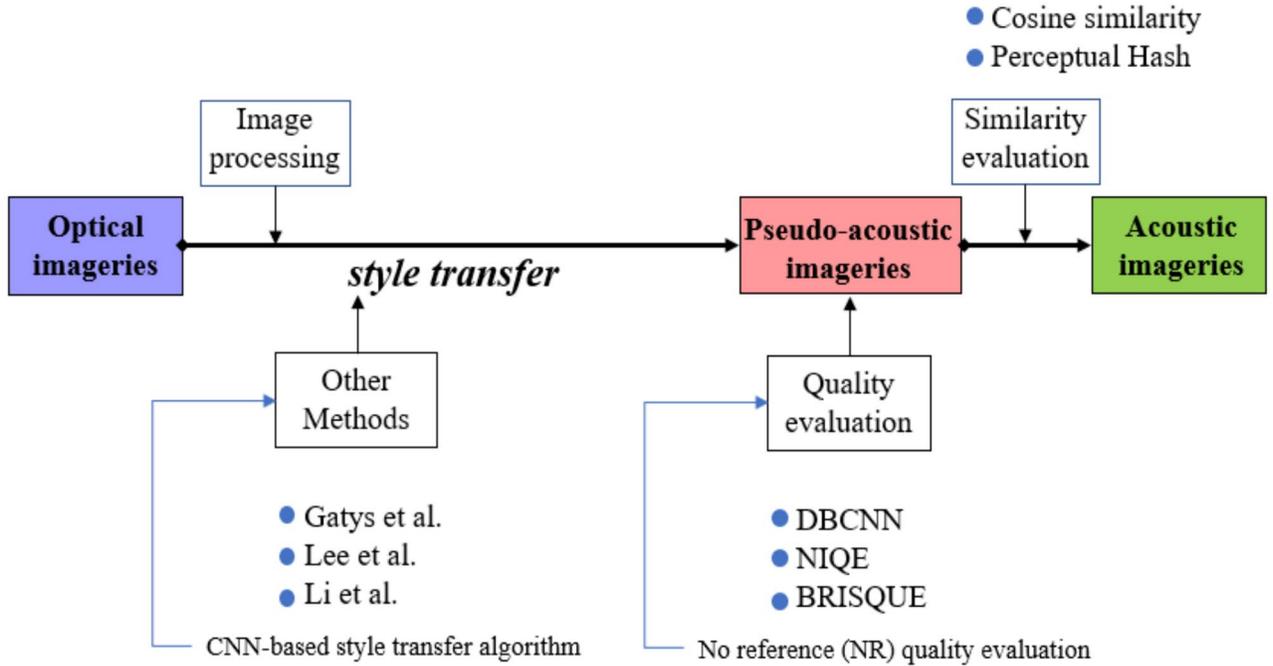

Figure 4. The flowchart of the overall experimental design.

### 3.1 Content image dataset

In the stage of training the style transfer model, this manuscript cites the marine fish dataset provided in the literature [19], which is rich in variety and contains variations in scale and orientation. It is initially used to train the image classifier and segmentation model. This manuscript uses this marine fish dataset to train a stylistic transfer model, mainly considering that the visual presentation of these samples is very close to the bird's eye view style presentation of sonar, which facilitates the rendering of optical images later using the transfer model. The model treats the marine fish dataset as image content for input, and the number of training samples collected is 9000, and some of the sample images are displayed as follows.

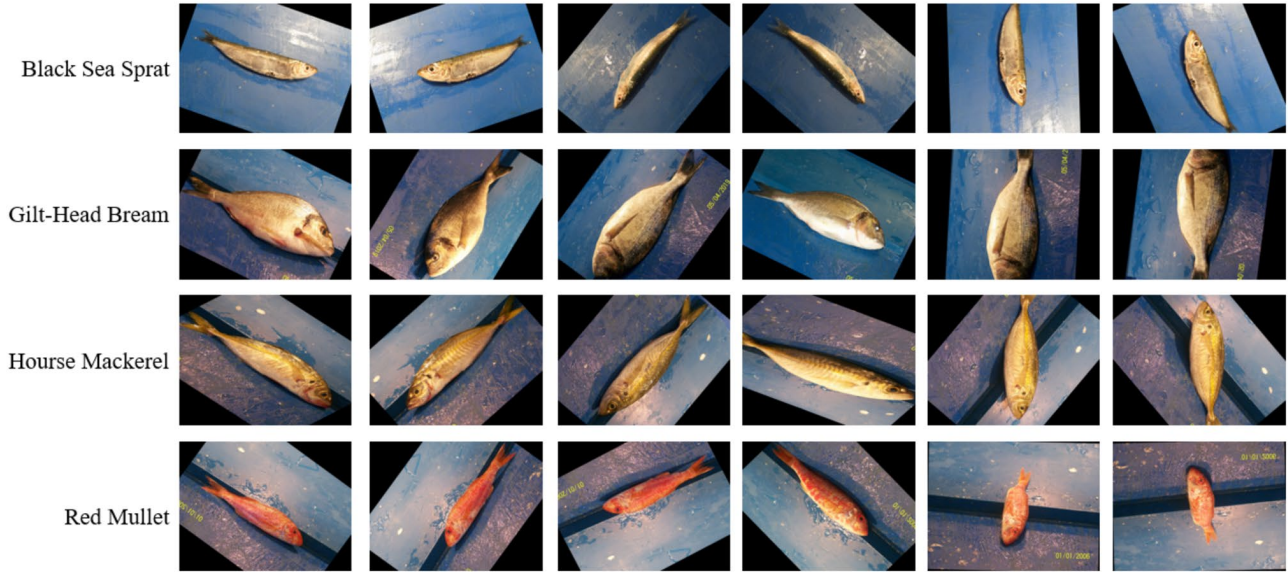

Figure 5. Examples of marine fish samples.

### 3.2 Style image dataset

The style dataset of the input model is then sourced from the image gallery of SOUND METRICS [4]. To ensure the model's generalization ability, this manuscript does not classify the acoustic imaging styles because different ocean backgrounds will produce different acoustic image styles depending on the specific application scenario. This manuscript performs frame decomposition on the images in GIF format in the dataset and finally collects 3952 training samples, some of which represent image styles as shown in Figure 6.

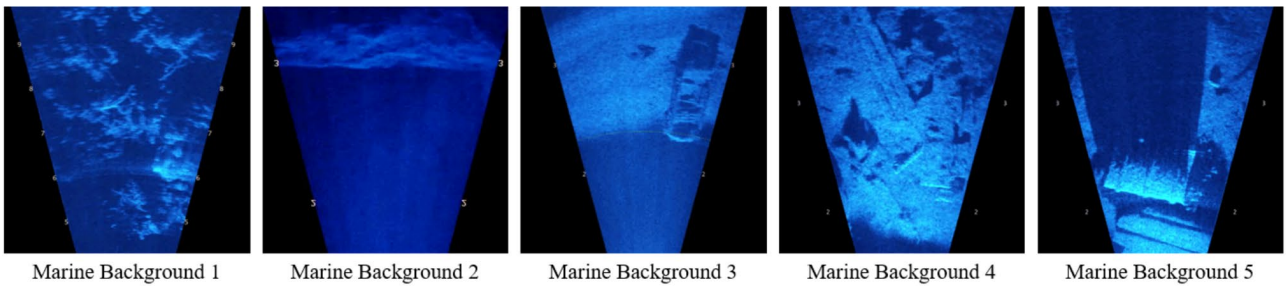

Figure 6. Examples of DIDSON imagery data.

### 3.3 Verification method

After completing the training of the style transfer model, this manuscript analyzes the transfer effects from qualitative and quantitative perspectives, in which DBCNN [20], NIQE [21], and BRISQUE [22] are introduced to evaluate the own quality of the generated pseudo-acoustic samples quantitatively. Cosine similarity [23] and Perceptual Hash [24] are used to quantitatively evaluate the similarity between the generated pseudo-acoustic samples and the authentic acoustic images. In addition, a comprehensive comparison between the method in this paper and the style transfer methods used in [10], [14], and [15] is also performed, and the comparison results are presented in schematic and tabular form.

### 3.4 Testing environment

All experiments were performed on a 64-bit PC with a Windows 10 operating system equipped with an Intel (R) Xeon (R) Gold 6226R 2.90 GHz processor, 128 GB of physical memory, and an NVIDIA Tesla T4 graphics card. Python 3.8 was used to compile the program.

## 4. EXPERIMENTAL RESULTS AND EVALUATION

### 4.1 Comparison with other transfer methods

In this subsection, the rendering effects of this paper are first compared with other typical transfer methods, and three pseudo-acoustic image samples are randomly selected for demonstration, considering the limitation of manuscript length. The comprehensive comparison results are shown in Figure 7 and Table 1, Table 2, and Table 3.

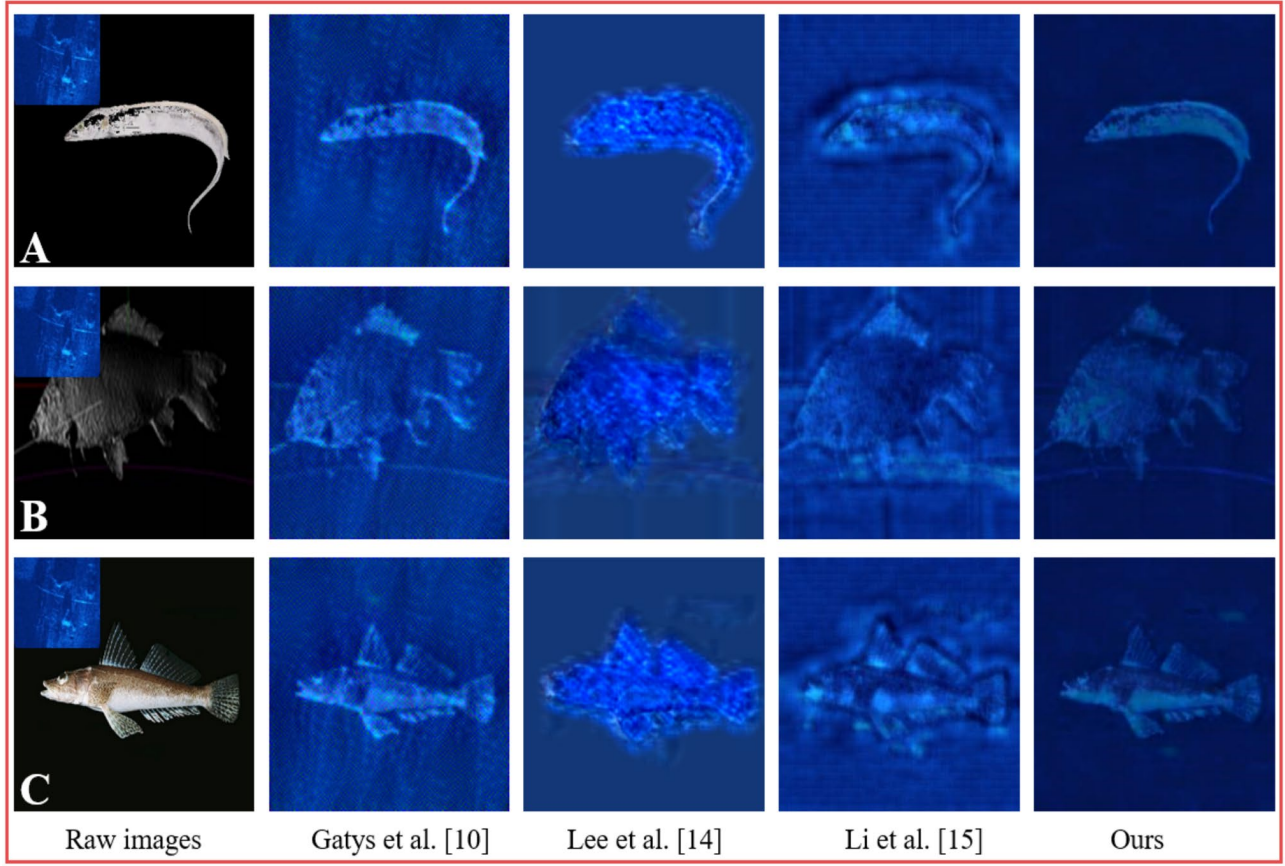

Figure 7. The transferred effects diagram corresponding to different methods.

Table 1. Quality assessment results of pseudo-acoustic sample A generated by different transfer methods.

| Methods / Metrics | DBCNN (higher - better) | NIQE (lower - better) | BRISQUE (lower - better) |
|---|---|---|---|
| Gatys et al. [10] | 29.3614 | 10.7025 | 91.2619 |
| Lee et al. [14] | 24.8291 | 10.3499 | 75.1481 |
| Li et al. [15] | 14.3595 | 7.3146 | 32.5592 |
| Ours | 25.0076 | 7.1349 | 62.7893 |

Table 2. Quality assessment results of pseudo-acoustic sample B generated by different transfer methods.

| Methods \ Metrics | DBCNN (higher - better) | NIQE (lower - better) | BRISQUE (lower - better) |
|---|---|---|---|
| Gatys et al. [10] | 28.7913 | 10.5685 | 90.9425 |
| Lee et al. [14] | 21.8231 | 9.8479 | 68.2902 |
| Li et al. [15] | 17.4809 | 6.9175 | 40.8156 |
| Ours | 22.9797 | 8.1695 | 62.7892 |

Table 3. Quality assessment results of pseudo-acoustic sample C generated by different transfer methods.

| Methods \ Metrics | DBCNN (higher - better) | NIQE (lower - better) | BRISQUE (lower - better) |
|---|---|---|---|
| Gatys et al. [10] | 25.4517 | 10.2652 | 83.1222 |
| Lee et al. [14] | 20.8625 | 7.6013 | 48.9831 |
| Li et al. [15] | 18.5751 | 7.2107 | 45.1564 |
| Ours | 22.0630 | 8.1983 | 58.2721 |

Figure 7 shows that the visual features of the pseudo-acoustic images generated by the transfer method in this paper are more apparent and more complete in structure, and the rendering effect is more in-depth. In contrast, other transfer methods will destroy the image details, and the rendering effect is unsatisfactory. From the statistical results in Table 1 to Table 3, it could be analyzed that the quality of the pseudo-acoustic images generated by the method in this letter has apparent advantages, which will be beneficial to the later training of the downstream task model.

### 4.2 Comparison of transfer effects in various acoustic scenes

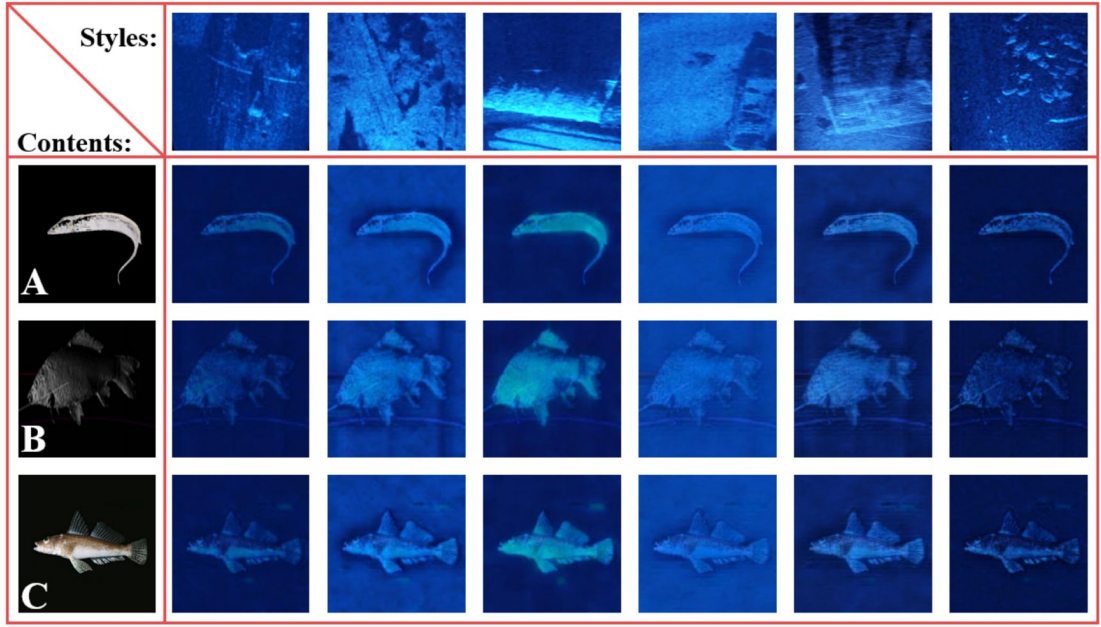

Figure 8. The effects of style transfer in various acoustic scenes.

This subsection demonstrates the rendering effects of the transfer method proposed in this letter in different underwater acoustic scenes. As seen in Figure 8, the method we proposed could be well adapted to various acoustic scenes, and the style texture features provided by these scenes also could be well learned and embedded. In practical underwater engineering applications, the operator could choose the transfer mode according to the acoustic environment of the specific underwater tasks and finally achieve the purpose of high-precision image datasets generation.

### 4.3 Comparison with real underwater acoustic images

In this subsection, the style transfer model proposed is used to transfer practical fish imageries. The final generated pseudo-acoustic images are evaluated for similarity with the natural acoustic camera-captured images [25] to verify the method's effectiveness in this paper. The comparison schematic is shown in Figure 9, and the similarity evaluation results are shown in Table 4.

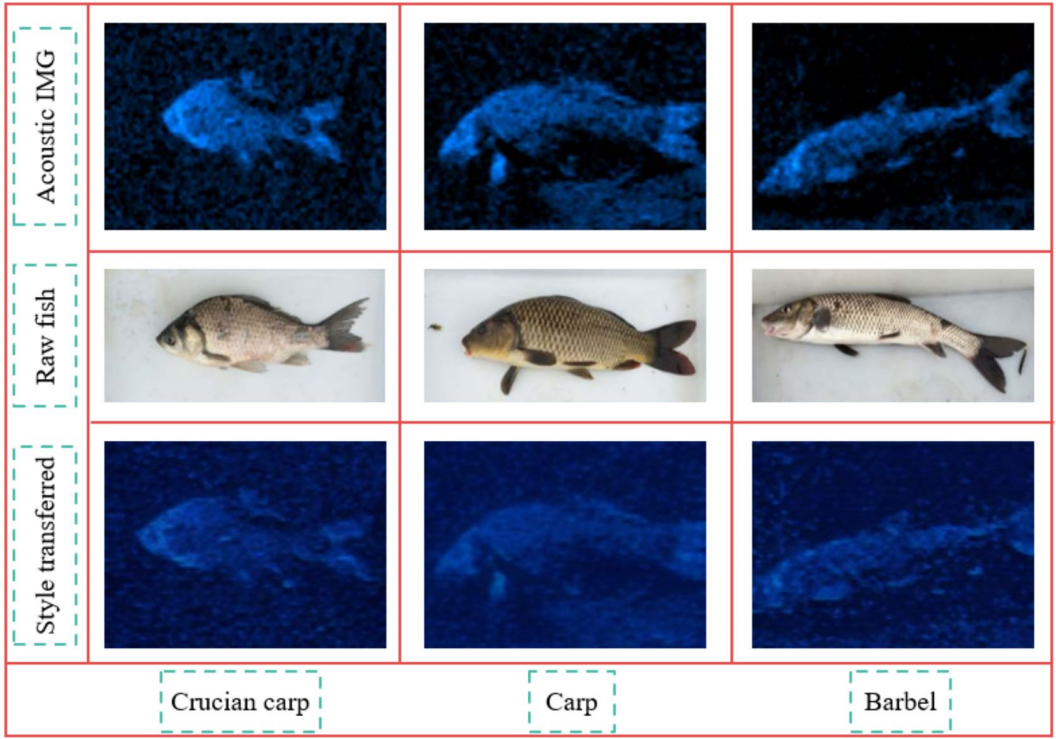

Figure 9. Comparison of real acoustic images of marine fish and the pseudo-acoustic images generated by style transfer.

Table 4. Similarity evaluation results between generated pseudo-acoustic images and real acoustic images.

| Fish<br>Metrics | Crucian carp | Carp | Barbel |
|---|---|---|---|
| Cosine similarity | 0.8290 | 0.6962 | 0.8388 |
| Perceptual Hash | 0.7031 | 0.8282 | 0.7187 |
| Average similarity | 0.7661 | 0.7622 | 0.7788 |

The experimental findings demonstrate that the approach suggested in this study could produce pseudo-acoustic images that are highly comparable to natural underwater acoustic imaging and realistic. It is important to note that, as illustrated in Figure 9, the pseudo-acoustic image produced by the method described in this letter still retains the noisy features of the image, which are an essential aspect of underwater acoustic imaging. Additionally, this validates the model's functionality from the side. In subsequent work, we intend to train a marine fish recognizer using the generated pseudo-acoustic pictures and further assess the effects of style transfer using the recognition's accuracy and error.

## 5. DISCUSSIONS

The purpose of acoustic-optical image style transfer is to make the feature distributions of the target and source domains closer, reduce the domain bias due to the difference in imaging mechanisms, and then transfer the knowledge information effectively between the two domains. From the experimental results, it can be seen that realistic underwater pseudo-acoustic samples can be generated using the transformer-based style transfer method. From a qualitative perspective, compared with other CNN-based methods, the visual effect of the pseudo-acoustic images generated using this paper is closer to the underwater acoustic detection effect, the image resolution is higher, and the details are better preserved. From the quantitative point of view, the generated pseudo-acoustic images have higher quality and higher similarity between them and authentic acoustic images, which can be directly used to train target recognition models. However, in the overall experimental process, the method proposed in this paper still has some shortcomings: (1) The style transfer method proposed in this paper can effectively retain image details and better affect the style transfer of acoustic images. However, when migrating the high-level semantic content of optical images, the transfer of feature details is too clear, resulting in the detail discrimination of the final generated pseudo-acoustic images exceeding the DIDSON device's discriminative ability. Later, we will add random noise to make the generated pseudo-acoustic images closer to the authentic captured underwater acoustic images. (2) The transformer-based style transfer method requires high computing resources. It does not have advantages in speed, which does not apply to portable computing devices at present, so the lightweight model will be one of our following research goals. (3) Due to the limitation of current experimental conditions, the sonar image dataset and optical image dataset selected in this paper are not strictly aligned in terms of semantic content. In the future, we will detect more acoustic-optical images of fish from natural ocean scenes and study the effect of unilateral domain transfer after aligning the semantic information of acoustic-optical images, which will open new horizons for matching and reconstruction tasks based on acoustic-optical images.

## 6. CONCLUSIONS

This paper presents a framework for generating underwater pseudo-acoustic sample datasets using a transformer-based style transfer method and it is tested on a natural marine fish dataset. The experimental results show that the quality of the pseudo-acoustic images generated by this model is better than the existing acoustic image transfer methods, and the average similarity between the pseudo-acoustic images generated by this model and the authentic marine acoustic images is close to 0.8. In the future, to validate the proposed framework in practical marine engineering applications, we will use this method to generate a sufficiently large dataset of underwater pseudo-acoustic images, and then, we will use this dataset to train an automated marine fish Class recognition model. In addition, we will simplify the framework to adapt it to the software equipment of AUVs to enhance the autonomous underwater operation capabilities of AUVs.

## ACKNOWLEDGMENT

This work was supported by the Chinese Shandong Provincial Key Research and Development Plan, under Grant No. 2019GHZ011 and No. 2021CXGC010702.

## REFERENCES

[1] Blondel, Philippe, and Bramley J. Murton. "Handbook of seafloor sonar imagery,". Vol. 7. Chichester: Wiley, 1997.
[2] Hurtos, Natalia, David Ribas, Xavier Cufi, Yvan Petillot, and Joaquim Salvi. "Fourier‐based registration for robust forward‐looking sonar mosaicing in low‐visibility underwater environments," Journal of Field Robotics 32(1), 123-151 (2015).
[3] Negahdaripour, Shahriar, "On 3-D motion estimation from feature tracks in 2-D FS sonar video." IEEE Transactions on Robotics 29(4), 1016-1030 (2013).
[4] "Sound Metrics,"<http://www.soundmetrics.com/> (6 September 2022).
[5] Teng, Bowen, and Hongjian Zhao, "Underwater target recognition methods based on the framework of deep learning: A survey," International Journal of Advanced Robotic Systems 17(6), (2020).
[6] Krizhevsky, Alex, Ilya Sutskever, and Geoffrey E. Hinton, "Imagenet classification with deep convolutional neural networks," Communications of the ACM 60(6), 84-90 (2017).

[7] O'Mahony, Niall, Sean Campbell, Anderson Carvalho, Suman Harapanahalli, Gustavo Velasco Hernandez, Lenka Krpalkova, Daniel Riordan, and Joseph Walsh, "Deep learning vs. traditional computer vision," In Science and information conference. Springer. Cham, 128-144 (2019).

[8] Zhuang, Fuzhen, Zhiyuan Qi, Keyu Duan, Dongbo Xi, Yongchun Zhu, Hengshu Zhu, Hui Xiong, and Qing He, "A comprehensive survey on transfer learning," Proceedings of the IEEE 109(1) 43-76 (2020).

[9] Quinonero-Candela, Joaquin, Masashi Sugiyama, Anton Schwaighofer, and Neil D. Lawrence, "Dataset shift in machine learning," Mit Press, (2022).

[10] Gatys, Leon A., Alexander S. Ecker, and Matthias Bethge, "Image style transfer using convolutional neural networks," In Proceedings of the IEEE conference on computer vision and pattern recognition, 2414-2423 (2016).

[11] Sung, Minsung, Jason Kim, Meungsuk Lee, Byeongjin Kim, Taesik Kim, Juhwan Kim, and Son-Cheol Yu, "Realistic sonar image simulation using deep learning for underwater object detection," International Journal of Control, Automation and Systems 18(3), 523-534 (2020).

[12] Liu, Dingyu, Yusheng Wang, Yonghoon Ji, Hiroshi Tsuchiya, Atsushi Yamashita, and Hajime Asama, "CycleGAN-based realistic image dataset generation for forward-looking sonar," Advanced Robotics 35(3)-(4), 242-254 (2021).

[13] Rixon Fuchs, Louise, Christer Larsson, and Andreas Gällström, "Deep learning based technique for enhanced sonar imaging," In 5th Underwater Acoustics Conference & Exhibition (UACE), Hersonissos, Crete, Greece, Jun. 30 2019-Jul. 5 2019, 1021-1028 (2019).

[14] Lee, Sejin, Byungjae Park, and Ayoung Kim, "Deep learning based object detection via style-transferred underwater sonar images," IFAC-PapersOnLine 52(21), 152-155 (2019).

[15] Li, Chuanlong, Xiufen Ye, Dongxiang Cao, Jie Hou, and Haibo Yang, "Zero shot objects classification method of side scan sonar image based on synthesis of pseudo samples," Applied Acoustics 173 (2021): 107691.

[16] Zhang, Y., K. Mizuno, A. Asada, Y. Fujimoto, and T. Shimada, "New method of fish classification by using high-resolution acoustic video camera-ARIS and local invariant feature descriptor," In Proc. of OCEANS, vol. 16. (2016).

[17] Schneider, Stefan, and Alex Zhuang, "Counting fish and dolphins in sonar images using deep learning," arXiv preprint arXiv:2007.12808 (2020).

[18] Deng, Yingying, Fan Tang, Weiming Dong, Chongyang Ma, Xingjia Pan, Lei Wang, and Changsheng Xu, "StyTr2: Image Style Transfer with Transformers," In Proceedings of the IEEE/CVF Conference on Computer Vision and Pattern Recognition, 11326-11336 (2022).

[19] Ulucan, Oguzhan, Diclehan Karakaya, and Mehmet Turkan, "A Large-Scale Dataset for Fish Segmentation and Classification," In 2020 Innovations in Intelligent Systems and Applications Conference (ASYU), IEEE, 1-5 (2020).

[20] Zhang, Weixia, Kede Ma, Jia Yan, Dexiang Deng, and Zhou Wang, "Blind image quality assessment using a deep bilinear convolutional neural network," IEEE Transactions on Circuits and Systems for Video Technology 30(1) 36-47 (2018).

[21] Mittal, Anish, Rajiv Soundararajan, and Alan C. Bovik, "Making a "completely blind" image quality analyzer," IEEE Signal processing letters 20(3), 209-212 (2012).

[22] Mittal, Anish, Anush Krishna Moorthy, and Alan Conrad Bovik, "No-reference image quality assessment in the spatial domain," IEEE Transactions on image processing 21(12), 4695-4708 (2012).

[23] Liu, Chengjun, "Discriminant analysis and similarity measure," Pattern Recognition 47(1), 359-367 (2014).

[24] Shuo-zhong, Wang, and Zhang Xin-peng, "Recent development of perceptual image hashing," Journal of Shanghai University (English Edition) 11(4), 323-331 (2007).

[25] Yu, Zhang, Katsunori Mizuno, Akira Asada, Yasufumi Fujimoto, and Tetsuo Shimada, "New method of fish classification by using high-resolution acoustic video camera-ARIS and local invariant feature descriptor," In OCEANS 2016 MTS/IEEE Monterey. IEEE, 1-6 (2016).